\documentclass{article}
\usepackage{base}

\usepackage{subcaption}
\usepackage{multirow}

\usepackage{hyperref}
\hypersetup{
    colorlinks=true,
    allcolors=blue,
}
\usepackage{url}
\usepackage{cleveref}
\usepackage{booktabs}

\title{Quantified Task Misalignment to Inform PEFT: An Exploration of Domain Generalization and Catastrophic Forgetting in CLIP}

\author{Laura Niss\thanks{Equal contribution} \and Kevin Vogt-Lowell$^*$ \and Theodoros Tsiligkaridis}
\date{MIT Lincoln Laboratory \\
\{laura.niss, kevin.vogt-lowell, ttsili\}@ll.mit.edu}

\begin{document}

\maketitle

\begin{abstract}
Foundations models are presented as generalists that often perform well over a myriad of tasks. Fine-tuning these models, even on limited data, provides an additional boost in task-specific performance but often at the cost of their wider generalization, an effect termed catastrophic forgetting. In this paper, we analyze the relation between task difficulty in the CLIP model and the performance of several simple parameter-efficient fine-tuning methods through the lens of domain generalization and catastrophic forgetting. We provide evidence that the silhouette score of the zero-shot image and text embeddings is a better measure of task difficulty than the average cosine similarity of correct image/label embeddings, and discuss observable relationships between task difficulty, fine-tuning method, domain generalization, and catastrophic forgetting. Additionally, the averaged results across tasks and performance measures demonstrate that a simplified method that trains only a subset of attention weights, which we call A-CLIP, yields a balance between domain generalization and catastrophic forgetting.
\end{abstract}

\section{Introduction}
\textbf{Motivation}
CLIP \citep{radfordLearningTransferableVisual2021a} has shown high performance on average over many disparate tasks, yet many applications still predominantly use it for specific downstream tasks that require further fine-tuning to obtain optimal performance. When fine-tuning the original model weights, varying degrees of catastrophic forgetting often occur, partially due to changes in the alignment between the text and image embedding spaces. This misalignment causes a decrease in model performance \citep{dingDonStopLearning2022, niContinualVisionLanguageRepresentation2023}. Several parameter-efficient fine-tuning (PEFT) methods have been shown to improve either in-domain or cross-dataset performance \citep{tourvon2022,liao2023descriptor}. With regard to multi-modal models, we speculate that there exists a relation between task difficulty, catastrophic forgetting, and domain generalization when considering such methods of fine-tuning. Methods that attenuate catastrophic forgetting could result in poor generalization if the method is too `surgical' in its learning, whereas methods that produce strong generalization on a specific task may distort other parts of the embedding space and trigger catastrophic forgetting. Motivated by this hypothesis, we consider five training datasets with two testing schema through which the performance and embedding space shifts of four PEFT methods were analyzed. To measure shifts in the embedding space and quantify task difficulty in terms of zero-shot alignment, we consider both the average cosine similarity between image and correct caption embeddings and the silhouette score, a measure of overlap between two clusters, between all image and text embeddings. 

\textbf{Contributions}
To the best of our knowledge, we are the first to provide evidence that, with respect to CLIP: 
1) the silhouette score of the zero-shot text and image embeddings can be used as a measure of `difficulty' of the target task in terms of how much the relevant portions of the image and text embedding spaces must shift to better align for the training task; 
2) BitFit is susceptible to catastrophic forgetting when the target task has a large zero-shot silhouette score; and
3) LoRA can significantly minimize catastrophic forgetting by restricting most changes in the embedding space to the area associated with the target task, though this precision can hamper performance gains relative to other methods in terms of domain generalization and in-domain accuracy.  

Additionally, we observe that our attention-based fine-tuning method, A-CLIP (a further refinement of \cite{tourvon2022}), yields the best balance of attenuating catastrophic forgetting while maintaining or improving performance in terms of domain generalization.

\section{Related Work}
Catastrophic forgetting in neural networks describes the significant drop in performance on previous tasks that accompanies increasing amounts of training. While this loss of knowledge has been proven to significantly affect traditional supervised learning techniques \citep{mccloskeyCatastrophicInterferenceConnectionist1989, kirkpatrickOvercomingCatastrophicForgetting2017}, recent research has shown that self-supervised, uni-modal models do not suffer from such severe forgetting \citep{ni2021selfsupervised, hu2022does}. Yet, with multi-modal foundation models, such as CLIP, continuous training can cause misalignment of the image and text embeddings, an outcome highlighted and explored in several works \citep{dingDonStopLearning2022,linSpecialityVsGenerality2023,niContinualVisionLanguageRepresentation2023,NEURIPS2022_bd361197,fan2022unified}.

PEFT methods could be considered among the parameter-isolation techniques explored in some catastrophic forgetting literature, and as such have the potential to alleviate catastrophic forgetting \citep{aljundi2018memory,laborieuxSynapticMetaplasticityBinarized2021,kirkpatrickOvercomingCatastrophicForgetting2017}. Motivated by the lack of research exploring the interplay between the misalignment problem and the possible attenuating effects of PEFT on catastrophic forgetting, we investigate the relationships between these fine-tuning methods, zero-shot (ZS) task alignment, domain generalization, and catastrophic forgetting.

\section{Methodology}
\label{sec:methods}


\textbf{Model Selection}
Our simple PEFT method, A-CLIP, freezes all of the weights in CLIP except for the in-projection weights of the attention layers, reducing the number of trainable weights even further than \cite{tourvon2022}. We chose this method based on observations that with some learning parameters, the in-projection weights of CLIP's attention layers changed significantly more than the out-projection layers during fine-tuning. Further discussion can be found in \Cref{append:aclip}.

In addition to A-CLIP, we selected three SOTA PEFT methods for comparison, including CLIP-Adapter \citep{gaoCLIPAdapterBetterVisionLanguage2023}, LoRA \citep{huLoRALowRankAdaptation2022a}, and BitFit (bias-tuning) \citep{benzakenBitFitSimpleParameterefficient2022}. All methods use CLIP ViT-B/32 provided by OpenCLIP \citep{Cherti_2023_CVPR} as the model backbone and were also compared to a standard CLIP ViT-B/32 model fine-tuned end-to-end (full) using cross-entropy loss.

\textbf{Alignment Measures}
We consider two measures of alignment: the average cosine similarity (cosine) of image embeddings and their correct caption embeddings, and the silhouette score (ss) between the image and text embeddings. The silhouette score represents an average measure of the distance among points in their own cluster compared to that of the next nearest neighbor. A value of 0 implies cluster overlap and a value of 1 implies cluster separation. Formulas for both measures are found in \Cref{append:align}.

\section{Results}
\label{sec:results}
\textbf{Datasets}
To investigate zero-shot task alignment and model performance across a variety of domain shift magnitudes, we train our models on five diverse datasets divided into two cross-dataset evaluation schema: one with small domain shifts to evaluate domain generalization (DG) and another with large domain shifts to evaluate catastrophic forgetting (CF). For DG, we use two training sets: 1) 16-shot ImageNet \citep{dengImageNet2009} evaluated on ImageNet-V2, ImageNet-Sketch, ImageNet-Rendition, and ImageNet-Adversarial, and 2) the in-distribution subset of the Functional Map of the World (FMoW-ID) satellite dataset \citep{christie_functional_2018} from WILDS \citep{kohWILDSBenchmarkInTheWild2021} evaluated on the satellite imagery benchmarks FMoW-OOD, RESISC45, and EuroSAT. For CF, we selected three datasets on which ZS CLIP has exhibited extreme performances: 1) Stanford Cars \citep{krauseStanfordCars}, on which zero-shot CLIP outperforms ResNet with a linear probe,  2) GTSRB \citep{stallcampGTSRB2012} which underperforms ResNet with a linear probe, and 3) SVHN \citep{netzerReadingDigits2011} on which zero-shot CLIP also performs quite poorly. These datasets are evaluated with additional common vision benchmarks, namely CIFAR100 \citep{krizhevskyCIFAR100}, DTD \citep{cimpoi14DTDa}, MNIST \citep{lecun1998MNIST}, STL10 \citep{coates2011STL10}, and SUN397 \citep{xiao2010SUN397}.

\textbf{Experimental Setting}
Training for each method involved 40 epochs of learning using mini-batch stochastic gradient descent. The model checkpoint with the highest validation accuracy during training was used as the final model, and reported accuracies were calculated using held-out test data. Given that our interest lies in comparing models across task settings and evaluating their general performance rather than comparing state-of-the-art performances between PEFT methods, we opted to borrow several hyper-parameter selections made in \cite{liao2023descriptor}. Training details for each model can be found in \Cref{append:methods}.

\begin{figure}[t]
    \centering
    \includegraphics[width=0.45\linewidth]{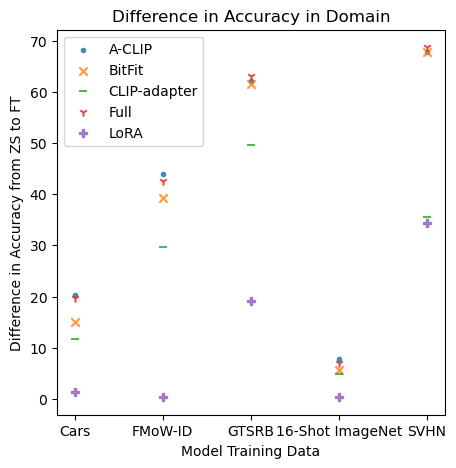}
    \includegraphics[width=0.45\linewidth]{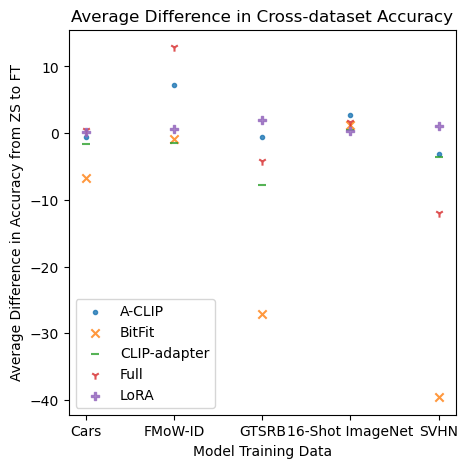}
    \caption{ In-domain and cross-dataset accuracy by model and training data.}
    \label{fig:acc}
\end{figure}

\begin{figure}[t]
    \centering
\includegraphics[width=.6\linewidth]{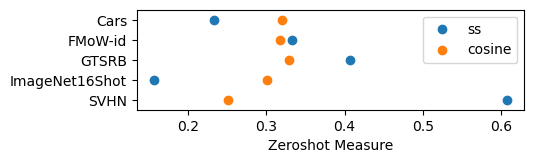}
    \caption{Silhouette score and average cosine similarity of zero-shot image and text embeddings.}
    \label{fig:silhouette}
\end{figure}

\textbf{Model Performance to Evaluation Task} In \Cref{fig:acc}, both plots show changes in accuracy relative to that of ZS CLIP. The in-domain (ID) accuracy, seen in the left-hand plot, shows that across training and testing schema, full fine-tuning, A-CLIP, and BitFit achieved the highest accuracies, LoRA the lowest, and CLIP-Adapter varying in between. It is interesting to note that the general shape of this plot maps well to the ZS silhouette score seen in \Cref{fig:silhouette}. We interpret this similarity as the silhouette score (SS) providing some information about how much more can be learned. If the SS is large, then there exists a significant change in alignment that can happen during training that may improve performance, whereas when the SS is small, the embeddings are already well-aligned and there may be relatively less gain in performance.

The right-hand plot of \Cref{fig:acc} shows the performance for our DG and CF evaluation schema. We see that LoRA does not appear to suffer from CF, though analysis of alignment changes suggests models with better ID performance would induce CF (\Cref{append:measure change}). In contrast, A-CLIP generally maintains ZS CLIP's performance on the CF evaluations and improves in its DG performance as seen by the FMoW-ID and 16-shot ImageNet results. Full fine-tuning does well for DG but is subject to some CF, as seen in the evaluations on GTSRB and SVHN. We see similar results with CLIP-Adapter. BitFit, while obtaining good ID performance, suffers from the greatest amount of CF and shows either no improvement or marginal loss for DG when compared to the other methods. Again, we see the loss in performance on the three CF evaluations mirrors the SS pattern in \Cref{fig:silhouette}.

\begin{figure}
    \centering
    \includegraphics[width=0.45\linewidth]{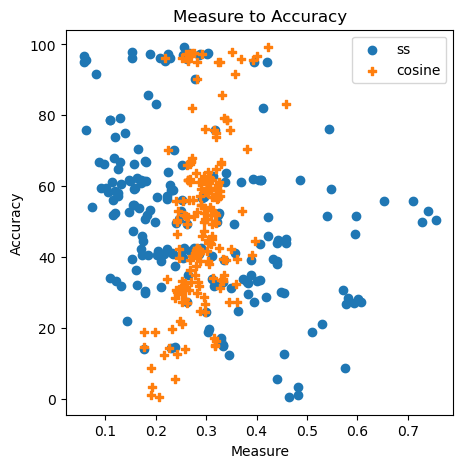}
\includegraphics[width=0.45\linewidth]{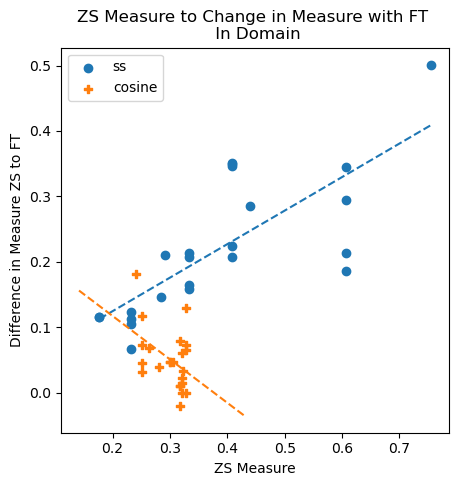}

    \includegraphics[width=0.4\linewidth]{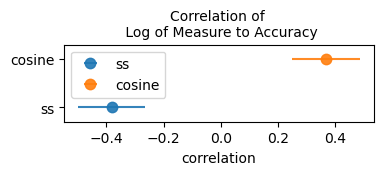}
\includegraphics[width=0.4\linewidth]{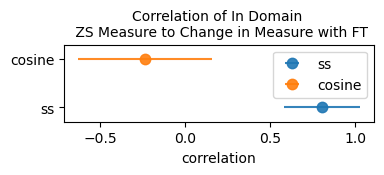}
    \caption{The top left figure compares alignment measure to accuracy and contains all training methods and test data. The bottom plots shows the Pearson correlation with 95\% confidence interval, with the left using the log of the measures given that their relationship to accuracy appears exponential. The top right figure contains only ID test data and excludes results from 16-shot ImageNet and LoRA.}
    \label{fig:measure to acc}
\end{figure}

\textbf{Silhouette Score, Accuracy, and Difficulty} To further test whether the SS holds empirically as an approximate measure of difficulty, we compare the SS and average cosine similarity (ACS) to accuracy. In \Cref{fig:measure to acc}, the top left plot shows all results, ZS and fine-tuned (FT). The Pearson correlation with 95\% confidence intervals in the bottom left plot uses the log of the measure, as the relationship appears exponential. As expected, the ACS has a positive correlation and the SS has a negative correlation, both with moderate strength. 

The top right plot of \Cref{fig:measure to acc} shows the relation of the ZS measure to the change in measure from fine-tuning for all ID results, excluding results trained on 16-shot ImageNet or with LoRA. These exclusions were made because few-shot learning is an inherently different setting from the other tasks, and because LoRA trained poorly on many of the ID tasks. We included six additional fully fine-tuned models from six datasets for more power: CIFAR100, DTD, EuroSAT, MNIST, RESISC45, and SUN397. The change in SS and the ZS measure are strongly correlated and display a smaller variance than that of the ACS. The slope of the SS regression line is statistically significant with a p-value of 0.001, while that of ACS is not significant with a p-value of 0.071. We conclude that, compared to the ACS, the SS provides more information about the `difficulty' of a task in terms of how misaligned its ZS embeddings are, and equally informs how much `room for improvement' remains while having a similarly strong relationship as ACS to accuracy.

\section{Conclusion}
In this paper, we gave evidence using CLIP that the silhouette score can be used as an approximation of `difficulty' in terms of how misaligned a task is and equally how much `room for improvement' remains. We also provided evidence that several simple and well-known fine-tuning methods have different desirable and undesirable traits that can be exacerbated by the difficultly of the task. In particular, we highlight BitFit's susceptibility to catastrophic forgetting and A-CLIP's balanced performance for both ID testing and domain generalization. We believe further work can be done to quantify the alignment of ZS CLIP and other multi-modal models to inform fine-tuning strategies based on desired outcomes as well as similarities between tasks.

\section*{Acknowledgements}
The authors acknowledge 
the MIT Lincoln Laboratory Supercomputing Center 
for providing the high performance computing resources that have contributed to the research results reported within this paper.

DISTRIBUTION STATEMENT A. Approved for public release. Distribution is unlimited. This material is based upon work supported by the Under Secretary of Defense for Research and Engineering under Air Force Contract No. FA8702-15-D-0001. Any opinions, findings, conclusions or recommendations expressed in this material are those of the author(s) and do not necessarily reflect the views of the Under Secretary of Defense for Research and Engineering.

© 2024 Massachusetts Institute of Technology.

Delivered to the U.S. Government with Unlimited Rights, as defined in DFARS Part 252.227-7013 or 7014 (Feb 2014). Notwithstanding any copyright notice, U.S. Government rights in this work are defined by DFARS 252.227-7013 or DFARS 252.227-7014 as detailed above. Use of this work other than as specifically authorized by the U.S. Government may violate any copyrights that exist in this work.

\clearpage

\bibliography{bib}

\begin{thebibliography}{29}
\providecommand{\natexlab}[1]{#1}
\providecommand{\url}[1]{\texttt{#1}}
\expandafter\ifx\csname urlstyle\endcsname\relax
  \providecommand{\doi}[1]{doi: #1}\else
  \providecommand{\doi}{doi: \begingroup \urlstyle{rm}\Url}\fi

\bibitem[Radford et~al.(2021)Radford, Kim, Hallacy, Ramesh, Goh, Agarwal, Sastry, Askell, Mishkin, Clark, Krueger, and Sutskever]{radfordLearningTransferableVisual2021a}
Alec Radford, Jong~Wook Kim, Chris Hallacy, Aditya Ramesh, Gabriel Goh, Sandhini Agarwal, Girish Sastry, Amanda Askell, Pamela Mishkin, Jack Clark, Gretchen Krueger, and Ilya Sutskever.
\newblock Learning {{Transferable Visual Models From Natural Language Supervision}}.
\newblock In \emph{Proceedings of the 38th {{International Conference}} on {{Machine Learning}}}, pages 8748--8763. {PMLR}, July 2021.

\bibitem[Ding et~al.(2022)Ding, Liu, Tian, Yang, and Ding]{dingDonStopLearning2022}
Yuxuan Ding, Lingqiao Liu, Chunna Tian, Jingyuan Yang, and Haoxuan Ding.
\newblock Don't {{Stop Learning}}: {{Towards Continual Learning}} for the {{CLIP Model}}, July 2022.
\newblock URL \url{http://arxiv.org/abs/2207.09248}.

\bibitem[Ni et~al.(2023)Ni, Wei, Tang, Zhuang, and Tian]{niContinualVisionLanguageRepresentation2023}
Zixuan Ni, Longhui Wei, Siliang Tang, Yueting Zhuang, and Qi~Tian.
\newblock Continual {{Vision-Language Representation Learning}} with {{Off-Diagonal Information}}, June 2023.
\newblock URL \url{https://arxiv.org/abs/2305.07437}.

\bibitem[Touvron et~al.(2022)Touvron, Cord, El-Nouby, Verbeek, and J\'{e}gou]{tourvon2022}
Hugo Touvron, Matthieu Cord, Alaaeldin El-Nouby, Jakob Verbeek, and Herv\'{e} J\'{e}gou.
\newblock Three things everyone should know about vision transformers.
\newblock In \emph{Computer Vision – ECCV 2022: 17th European Conference, Tel Aviv, Israel, October 23–27, 2022, Proceedings, Part XXIV}, page 497–515, Berlin, Heidelberg, 2022. Springer-Verlag.
\newblock ISBN 978-3-031-20052-6.
\newblock \doi{10.1007/978-3-031-20053-3_29}.
\newblock URL \url{https://doi.org/10.1007/978-3-031-20053-3_29}.

\bibitem[Liao et~al.(2023)Liao, Tsiligkaridis, and Kulis]{liao2023descriptor}
Christopher Liao, Theodoros Tsiligkaridis, and Brian Kulis.
\newblock Descriptor and word soups: Overcoming the parameter efficiency accuracy tradeoff for out-of-distribution few-shot learning, 2023.
\newblock URL \url{https://arxiv.org/pdf/2311.13612.pdf}.

\bibitem[McCloskey and Cohen(1989)]{mccloskeyCatastrophicInterferenceConnectionist1989}
Michael McCloskey and Neal~J. Cohen.
\newblock Catastrophic {{Interference}} in {{Connectionist Networks}}: {{The Sequential Learning Problem}}.
\newblock In \emph{Psychology of {{Learning}} and {{Motivation}}}, volume~24, pages 109--165. {Elsevier}, 1989.
\newblock ISBN 978-0-12-543324-2.
\newblock \doi{10.1016/S0079-7421(08)60536-8}.

\bibitem[Kirkpatrick et~al.(2017)Kirkpatrick, Pascanu, Rabinowitz, Veness, Desjardins, Rusu, Milan, Quan, Ramalho, {Grabska-Barwinska}, Hassabis, Clopath, Kumaran, and Hadsell]{kirkpatrickOvercomingCatastrophicForgetting2017}
James Kirkpatrick, Razvan Pascanu, Neil Rabinowitz, Joel Veness, Guillaume Desjardins, Andrei~A. Rusu, Kieran Milan, John Quan, Tiago Ramalho, Agnieszka {Grabska-Barwinska}, Demis Hassabis, Claudia Clopath, Dharshan Kumaran, and Raia Hadsell.
\newblock Overcoming catastrophic forgetting in neural networks.
\newblock \emph{Proceedings of the National Academy of Sciences}, 114\penalty0 (13):\penalty0 3521--3526, March 2017.
\newblock ISSN 0027-8424, 1091-6490.
\newblock \doi{10.1073/pnas.1611835114}.

\bibitem[Ni et~al.(2021)Ni, Tang, and Zhuang]{ni2021selfsupervised}
Zixuan Ni, Siliang Tang, and Yueting Zhuang.
\newblock Self-supervised class incremental learning, 2021.
\newblock URL \url{https://arxiv.org/abs/2111.11208}.

\bibitem[Hu et~al.(2022{\natexlab{a}})Hu, Yan, Lu, Hong, Hu, Zhang, Li, Wang, and Feng]{hu2022does}
Dapeng Hu, Shipeng Yan, Qizhengqiu Lu, Lanqing Hong, Hailin Hu, Yifan Zhang, Zhenguo Li, Xinchao Wang, and Jiashi Feng.
\newblock How well does self-supervised pre-training perform with streaming data?, 2022{\natexlab{a}}.
\newblock URL \url{https://arxiv.org/abs/2104.12081}.

\bibitem[Lin et~al.(2023)Lin, Tan, Lin, Zheng, Pi, Zhang, Diao, Wang, Zhao, Yao, and Zhang]{linSpecialityVsGenerality2023}
Yong Lin, Lu~Tan, Hangyu Lin, Zeming Zheng, Renjie Pi, Jipeng Zhang, Shizhe Diao, Haoxiang Wang, Han Zhao, Yuan Yao, and Tong Zhang.
\newblock Speciality vs {{Generality}}: {{An Empirical Study}} on {{Catastrophic Forgetting}} in {{Fine-tuning Foundation Models}}, October 2023.
\newblock URL \url{https://arxiv.org/abs/2309.06256}.

\bibitem[Srinivasan et~al.(2022)Srinivasan, Chang, Pinto~Alva, Chochlakis, Rostami, and Thomason]{NEURIPS2022_bd361197}
Tejas Srinivasan, Ting-Yun Chang, Leticia Pinto~Alva, Georgios Chochlakis, Mohammad Rostami, and Jesse Thomason.
\newblock {{CLiMB}}: {{A}} continual learning benchmark for vision-and-language tasks.
\newblock In S.~Koyejo, S.~Mohamed, A.~Agarwal, D.~Belgrave, K.~Cho, and A.~Oh, editors, \emph{Advances in Neural Information Processing Systems}, volume~35, pages 29440--29453. {Curran Associates, Inc.}, 2022.

\bibitem[Fan et~al.(2022)Fan, Wei, Chen, Wang, Li, Xu, and Huang]{fan2022unified}
Zhihao Fan, Zhongyu Wei, Jingjing Chen, Siyuan Wang, Zejun Li, Jiarong Xu, and Xuanjing Huang.
\newblock A unified continuous learning framework for multi-modal knowledge discovery and pre-training, 2022.
\newblock URL \url{https://arxiv.org/pdf/2206.05555.pdf}.

\bibitem[Aljundi et~al.(2018)Aljundi, Babiloni, Elhoseiny, Rohrbach, and Tuytelaars]{aljundi2018memory}
Rahaf Aljundi, Francesca Babiloni, Mohamed Elhoseiny, Marcus Rohrbach, and Tinne Tuytelaars.
\newblock Memory aware synapses: Learning what (not) to forget.
\newblock In \emph{Proceedings of the European conference on computer vision (ECCV)}, pages 139--154, 2018.

\bibitem[Laborieux et~al.(2021)Laborieux, Ernoult, Hirtzlin, and Querlioz]{laborieuxSynapticMetaplasticityBinarized2021}
Axel Laborieux, Maxence Ernoult, Tifenn Hirtzlin, and Damien Querlioz.
\newblock Synaptic metaplasticity in binarized neural networks.
\newblock \emph{Nature Communications}, 12\penalty0 (1):\penalty0 2549, May 2021.
\newblock ISSN 2041-1723.
\newblock \doi{10.1038/s41467-021-22768-y}.

\bibitem[Gao et~al.(2023)Gao, Geng, Zhang, Ma, Fang, Zhang, Li, and Qiao]{gaoCLIPAdapterBetterVisionLanguage2023}
Peng Gao, Shijie Geng, Renrui Zhang, Teli Ma, Rongyao Fang, Yongfeng Zhang, Hongsheng Li, and Yu~Qiao.
\newblock {{CLIP-Adapter}}: {{Better Vision-Language Models}} with {{Feature Adapters}}.
\newblock \emph{International Journal of Computer Vision}, 132\penalty0 (2):\penalty0 581--595, February 2023.
\newblock ISSN 0920-5691, 1573-1405.
\newblock \doi{10.1007/s11263-023-01891-x}.

\bibitem[Hu et~al.(2022{\natexlab{b}})Hu, Shen, Wallis, Allen{-}Zhu, Li, Wang, Wang, and Chen]{huLoRALowRankAdaptation2022a}
Edward~J. Hu, Yelong Shen, Phillip Wallis, Zeyuan Allen{-}Zhu, Yuanzhi Li, Shean Wang, Lu~Wang, and Weizhu Chen.
\newblock Lora: Low-rank adaptation of large language models.
\newblock In \emph{The Tenth International Conference on Learning Representations, {ICLR} 2022, Virtual Event, April 25-29, 2022}. OpenReview.net, 2022{\natexlab{b}}.
\newblock URL \url{https://openreview.net/forum?id=nZeVKeeFYf9}.

\bibitem[Ben~Zaken et~al.(2022)Ben~Zaken, Goldberg, and Ravfogel]{benzakenBitFitSimpleParameterefficient2022}
Elad Ben~Zaken, Yoav Goldberg, and Shauli Ravfogel.
\newblock {{BitFit}}: {{Simple Parameter-efficient Fine-tuning}} for {{Transformer-based Masked Language-models}}.
\newblock In Smaranda Muresan, Preslav Nakov, and Aline Villavicencio, editors, \emph{Proceedings of the 60th {{Annual Meeting}} of the {{Association}} for {{Computational Linguistics}} ({{Volume}} 2: {{Short Papers}})}, pages 1--9, {Dublin, Ireland}, May 2022. {Association for Computational Linguistics}.
\newblock \doi{10.18653/v1/2022.acl-short.1}.

\bibitem[Cherti et~al.(2023)Cherti, Beaumont, Wightman, Wortsman, Ilharco, Gordon, Schuhmann, Schmidt, and Jitsev]{Cherti_2023_CVPR}
Mehdi Cherti, Romain Beaumont, Ross Wightman, Mitchell Wortsman, Gabriel Ilharco, Cade Gordon, Christoph Schuhmann, Ludwig Schmidt, and Jenia Jitsev.
\newblock Reproducible scaling laws for contrastive language-image learning.
\newblock In \emph{Proceedings of the IEEE/CVF Conference on Computer Vision and Pattern Recognition (CVPR)}, pages 2818--2829, June 2023.

\bibitem[Deng et~al.(2009)Deng, Dong, Socher, Li, Li, and Fei-Fei]{dengImageNet2009}
Jia Deng, Wei Dong, Richard Socher, Li-Jia Li, Kai Li, and Li~Fei-Fei.
\newblock Imagenet: A large-scale hierarchical image database.
\newblock In \emph{2009 IEEE Conference on Computer Vision and Pattern Recognition}, pages 248--255, 2009.
\newblock \doi{10.1109/CVPR.2009.5206848}.

\bibitem[Christie et~al.(2018)Christie, Fendley, Wilson, and Mukherjee]{christie_functional_2018}
Gordon Christie, Neil Fendley, James Wilson, and Ryan Mukherjee.
\newblock Functional {Map} of the {World}.
\newblock In \emph{2018 {IEEE}/{CVF} {Conference} on {Computer} {Vision} and {Pattern} {Recognition}}, pages 6172--6180, Salt Lake City, UT, June 2018. IEEE.
\newblock ISBN 978-1-5386-6420-9.
\newblock \doi{10.1109/CVPR.2018.00646}.
\newblock URL \url{https://ieeexplore.ieee.org/document/8578744/}.

\bibitem[Koh et~al.(2021)Koh, Sagawa, Marklund, Xie, Zhang, Balsubramani, Hu, Yasunaga, Phillips, Gao, Lee, David, Stavness, Guo, Earnshaw, Haque, Beery, Leskovec, Kundaje, Pierson, Levine, Finn, and Liang]{kohWILDSBenchmarkInTheWild2021}
Pang~Wei Koh, Shiori Sagawa, Henrik Marklund, Sang~Michael Xie, Marvin Zhang, Akshay Balsubramani, Weihua Hu, Michihiro Yasunaga, Richard~Lanas Phillips, Irena Gao, Tony Lee, Etienne David, Ian Stavness, Wei Guo, Berton Earnshaw, Imran Haque, Sara~M Beery, Jure Leskovec, Anshul Kundaje, Emma Pierson, Sergey Levine, Chelsea Finn, and Percy Liang.
\newblock Wilds: A benchmark of in-the-wild distribution shifts.
\newblock In Marina Meila and Tong Zhang, editors, \emph{Proceedings of the 38th International Conference on Machine Learning}, volume 139 of \emph{Proceedings of Machine Learning Research}, pages 5637--5664. PMLR, 18--24 Jul 2021.
\newblock URL \url{https://proceedings.mlr.press/v139/koh21a.html}.

\bibitem[Krause et~al.(2013)Krause, Stark, Deng, and Fei-Fei]{krauseStanfordCars}
Jonathan Krause, Michael Stark, Jia Deng, and Li~Fei-Fei.
\newblock 3d object representations for fine-grained categorization.
\newblock In \emph{2013 IEEE International Conference on Computer Vision Workshops}, pages 554--561, 2013.
\newblock \doi{10.1109/ICCVW.2013.77}.

\bibitem[Stallkamp et~al.(2012)Stallkamp, Schlipsing, Salmen, and Igel]{stallcampGTSRB2012}
J.~Stallkamp, M.~Schlipsing, J.~Salmen, and C.~Igel.
\newblock Man vs. computer: Benchmarking machine learning algorithms for traffic sign recognition.
\newblock \emph{Neural Networks}, 32:\penalty0 323--332, 2012.
\newblock ISSN 0893-6080.
\newblock \doi{https://doi.org/10.1016/j.neunet.2012.02.016}.
\newblock URL \url{https://www.sciencedirect.com/science/article/pii/S0893608012000457}.
\newblock Selected Papers from IJCNN 2011.

\bibitem[Netzer et~al.(2011)Netzer, Wang, Coates, Bissacco, Wu, and Ng]{netzerReadingDigits2011}
Yuval Netzer, Tao Wang, Adam Coates, Alessandro Bissacco, Bo~Wu, and Andrew~Y. Ng.
\newblock Reading {Digits} in {Natural} {Images} with {Unsupervised} {Feature} {Learning}.
\newblock In \emph{{NIPS} {Workshop} on {Deep} {Learning} and {Unsupervised} {Feature} {Learning} 2011}, 2011.
\newblock URL \url{http://ufldl.stanford.edu/housenumbers/nips2011_housenumbers.pdf}.

\bibitem[Krizhevsky et~al.(2009)]{krizhevskyCIFAR100}
Alex Krizhevsky et~al.
\newblock Learning multiple layers of features from tiny images.
\newblock 2009.

\bibitem[Cimpoi et~al.(2014)Cimpoi, Maji, Kokkinos, Mohamed, , and Vedaldi]{cimpoi14DTDa}
M.~Cimpoi, S.~Maji, I.~Kokkinos, S.~Mohamed, , and A.~Vedaldi.
\newblock Describing textures in the wild.
\newblock In \emph{Proceedings of the {IEEE} Conf. on Computer Vision and Pattern Recognition ({CVPR})}, 2014.

\bibitem[LeCun et~al.(1998)LeCun, Bottou, Bengio, and Haffner]{lecun1998MNIST}
Yann LeCun, L{\'e}on Bottou, Yoshua Bengio, and Patrick Haffner.
\newblock Gradient-based learning applied to document recognition.
\newblock \emph{Proceedings of the IEEE}, 86\penalty0 (11):\penalty0 2278--2324, 1998.

\bibitem[Coates et~al.(2011)Coates, Ng, and Lee]{coates2011STL10}
Adam Coates, Andrew Ng, and Honglak Lee.
\newblock An analysis of single-layer networks in unsupervised feature learning.
\newblock In \emph{Proceedings of the fourteenth international conference on artificial intelligence and statistics}, pages 215--223. JMLR Workshop and Conference Proceedings, 2011.

\bibitem[{Xiao} et~al.(2010){Xiao}, {Hays}, {Ehinger}, {Oliva}, and {Torralba}]{xiao2010SUN397}
J.~{Xiao}, J.~{Hays}, K.~A. {Ehinger}, A.~{Oliva}, and A.~{Torralba}.
\newblock Sun database: Large-scale scene recognition from abbey to zoo.
\newblock In \emph{2010 IEEE Computer Society Conference on Computer Vision and Pattern Recognition}, pages 3485--3492, June 2010.
\newblock \doi{10.1109/CVPR.2010.5539970}.

\end{thebibliography}

\appendix
\section{Appendix}
\label{append}
\subsection{Method details}
\label{append:methods}
Training for each method involved 40 epochs of learning using mini-batch stochastic gradient descent with a momentum of 0.9 and a batch size of 128. A weight decay of 1e-5 was used for training on GTSRB, SVHN, and 16-shot ImageNet (and with the additional Full models used in \Cref{fig:measure to acc}), and a weight decay of 1e-4 was used for training on FMoW-ID. The learning rates employed per fine-tuning method can be seen in \Cref{tab:learning_rates}, with the only exception being that training on FMoW-ID via cross entropy used a learning rate of 1e-5 instead of 2e-5. No learning rate decay was applied.

\begin{table}[ht]

\centering
\caption{\label{tab:learning_rates} Learning rates per fine-tuning method}
\begin{tabular}{l|r}
    \toprule
    Method & Learning Rate \\
    \midrule
    Full & 2e-5 \\
    CLIP-Adapter & 6e-3 \\
    LoRA & 1e-5 \\
    BitFit & 1e-3 \\
    A-CLIP & 1e-5 \\
    \bottomrule
\end{tabular}

\end{table}

For CLIP-Adapter, we used a reduction of 4. LoRA used rank 4 for FMoW-ID and 16-Shot ImageNet and rank 16 for SVHN, GTSRB, and Cars. 

For 16-shot ImageNet, SVHN, and GTSRB, validation sets were generated by randomly selecting class-balanced samples from the training data. The validation set for FMoW-ID was provided by the WILDS collection.

All experiments were run non-distributed on a single NVIDIA V100 GPU using an Anaconda 2023b environment with CUDA 11.8.

\subsection{Results}
\label{append:results}
We include the raw results of the evaluations for reference. Though we bold the best performing model for each test, we remind the reader that we did not conduct a rigorous hyper-parameter search, therefore minuscule differences between model performances do not necessarily indicate superiority. Rather, we intend to highlight the large differences observed across testing schema. 

\Cref{tab:id results} shows the in-domain evaluation results, \Cref{tab:dg results} shows evaluation results under the domain generalization schema, and \Cref{tab:cf results} shows the results for the catastrophic forgetting schema. 
\begin{table}[ht]
    \centering
        \caption{In-domain test results.}
    \label{tab:id results}
    \begin{tabular}{l|c|c|c|c|c|c}
    \toprule
   \multirow{2}{*}{Data} & \multicolumn{6}{c}{ID Model Performance}\\ 
     & ZS 	& CLIP-Adapter & A-CLIP & BitFit	& Full & LoRA \\ \hline
       Cars &  58.87 & 70.60 & \textbf{79.12} & 73.98 & 78.66 & 60.30   \\ \hline
FMoW-ID &  14.78 & 44.43 & \textbf{58.67} & 53.98 & 57.30 & 15.24   \\ \hline
GTSRB &  33.65 & 83.28 & 95.68 & 95.19 & \textbf{96.72} & 52.87   \\ \hline
16-Shot ImageNet &  59.24 & 64.15 & \textbf{66.96} & 64.93 & 66.34 & 59.56   \\ \hline
SVHN &  27.27 & 62.78 & 95.16 & 95.15 & \textbf{96.00} & 61.70 \\ \bottomrule
    \end{tabular}
\end{table}
\begin{table}[ht]
    \centering
    \caption{Domain generalization test results.}
    \label{tab:dg results}
    \begin{tabular}{l|c|c|c|c|c|c}
    \toprule
   \multirow{2}{*}{Data} & \multicolumn{6}{c}{DG Model Performance}\\ 
     & ZS 	& CLIP-Adapter & A-CLIP & BitFit	& Full & LoRA \\ \hline
       FMoW-ID & 39.01 &37.51	&46.25	&38.10&	\textbf{51.89}	&39.61 \\ \hline
       16-Shot ImageNet & 48.06 & 48.53&	\textbf{50.74}	&49.35&	49.69&	48.37 \\ \bottomrule
    \end{tabular}
\end{table}
\begin{table}[ht]
    \caption{Catastrophic forgetting test results.}
    \label{tab:cf results}
    \centering
    \begin{tabular}{l|c|c|c|c|c|c}
    \toprule
   \multirow{2}{*}{Data} & \multicolumn{6}{c}{CF Model Performance}\\ 
     & ZS 	& CLIP-Adapter & A-CLIP & BitFit	& Full & LoRA \\ \hline
      Cars &52.24& 50.62	&51.64	&45.49	&\textbf{52.66}	&52.45 \\
      SVHN &56.19& 52.62&	53.03&	16.61&	44.27	&\textbf{57.25}\\
      GTSRB &55.4& 47.6&	54.75	&28.35&	51.23	&\textbf{57.37}\\
      \bottomrule
    \end{tabular}
\end{table}
\clearpage
\subsection{Alignment Measures}
\label{append:align}
The following are the definitions of the two alignment measure used in this paper:

\textbf{Average cosine similarity}

For image embeddings $\vi \in \sI$ and correct label embedding $\vt(\vi) \in \sT$, the average cosine similarity is 
\[
\text{ACS}(\sI,\sT) = \frac{1}{|\sI|} \sum_{\vi \in \sI} \frac{\vi \cdot \vt(\vi)}{||\vi||\cdot||\vt(i)||}
\]

\textbf{Silhouette score}

For a set of clusters $\{\sC_1, \sC_2..., \sC_K\}$, vector $\vi \in \sC_I$, and a distance metric $d$, let
\begin{align*}
    & a(\vi) = \frac{1}{|C_I|-1} \sum_{\vj \in C_I, \vj \neq \vi} d(\vi,\vj) \\
    & b(\vi) = \min_{J \neq I} \frac{1}{|C_{J}|} \sum_{\vj \in C_{J}} d(\vi,\vj) \\
    & s(\vi) = \frac{b(\vi) - a(\vi)}{\max\{b(\vi),a(\vi)\}}
\end{align*} 
The silhouette score is defined as the mean $s(\vi)$ over all points,
\[
ss = \frac{1}{\sum_{I=1}^K |\sC_I|} \sum_{\vi \in \cup_{I=1}^K \sC_I} s(\vi)
\]

We use the L2 norm as the distance metric in this work.

\subsection{A-CLIP}
\label{append:aclip}
\begin{figure}[ht]
    \centering
    \includegraphics[width=0.6\linewidth]{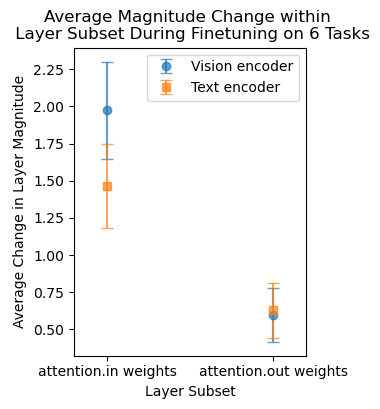}    
    \caption{Average difference of layer subset magnitude with 95\% confidence intervals. }
    \label{fig:ave diff}
\end{figure}

We trained six additional models with full fine-tuning on Cars, EuroSAT, SUN397, CIFAR100, SVHN, and GTSRB with a learning rate of 1e-5 and a weight decay of 0.05. This larger weight decay revealed a disparity in the magnitude of change in the attention in-projection and out-projection layers not seen with smaller weight decays, leading us to explore CLIP's performance when limiting training to only those weights in the attention layers. We coined this PEFT method A-CLIP. \Cref{fig:ave diff} shows the disparity of the magnitude changes of the two attention layers for the vision and text encoders. 

Though we refrain from a detailed discussion in this paper, we note that further empirical results on these datasets suggest that fine-tuning only the attention in-projection layers of the first few transformer blocks can achieve in-domain performance on par with full fine-tuning, and we suspect that such fine-tuning would maintain the DG and CF performance associated with training all of the attention in-projection layers as we have done with A-CLIP in this paper. 

\subsection{Alignment Measure Changes by Model}
\label{append:measure change}
\begin{figure}[ht]
    \centering
    \begin{subfigure}{\textwidth}
    \centering
        \includegraphics[width=0.45\textwidth]{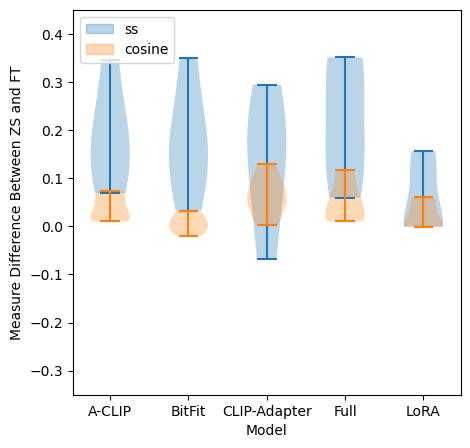}
        \caption{In-domain measure difference distribution by model.}
        \label{fig:model measure id}
    \end{subfigure}
    \begin{subfigure}{\textwidth}
        \centering
        \includegraphics[width=0.45\textwidth]{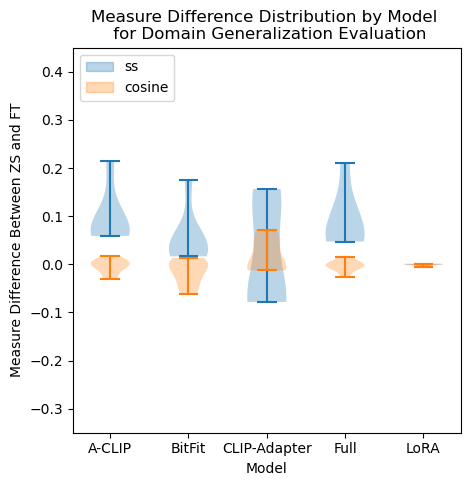}
    \includegraphics[width=0.45\textwidth]{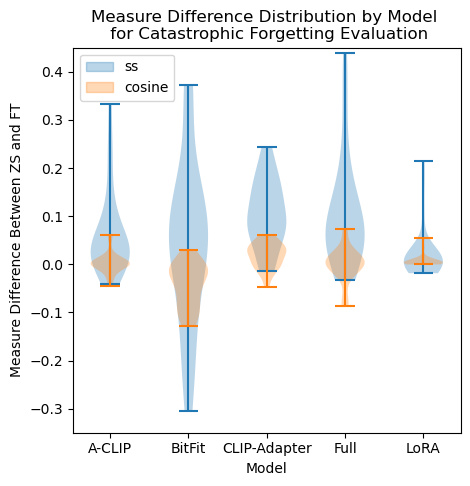}
    \caption{Measure difference by model for domain generalization and catastrophic forgetting.}
    \label{fig:model measure ood}
    \end{subfigure}
    \caption{Distributions of the difference in measure from the ZS to FT model for in-domain, DG, and CF. Here, `ss' is the silhouette score of the ZS text and image embeddings minus the silhouette score of the FT text and image embeddings. A positive value thus means the clusters of image and text embeddings moved closer together. Similarly, 'cosine' is the cosine similarity score of the FT text and image embeddings minus the cosine similarity of the ZS text and image embeddings. Again, a positive value means the image and appropriate label embeddings moved closer to each other.}
    \label{fig:model measure}
\end{figure}

Looking at both the average cosine similarity and silhouette score, we can compare the alignment changes imposed by each fine-tuning method. In \Cref{fig:model measure}, we compare the alignment changes in-domain, with respect to domain generalization, and with respect to catastrophic forgetting. Looking first at in-domain alignment in \Cref{fig:model measure id}, A-CLIP, BitFit, and Full have similar measure distributions, with Full having more weight for slightly stronger alignments. CLIP-Adapter and LoRA have measurably smaller changes in alignment, with LoRA being particularly and expectedly small based on its in-domain performance. 

In \Cref{fig:model measure ood}, there exist clear differences in the magnitude of measure change between models in both the domain generalization and catastrophic forgetting evaluation schema. In the domain generalization schema, LoRA has almost no change in alignment, yet it has clear changes in the catastrophic forgetting schema (though less relative to the other methods). These results suggest that hyper-parameter selections which produce better in-domain performance for LoRA will also likely exacerbate catastrophic forgetting. 

The other results align with our observations about evaluation performance discussed in \cref{sec:results}. We see BitFit inducing less change in alignment in the DG schema and significant changes in the CF schema. A-CLIP has similar changes to Full in the DG schema and less extreme changes than Full in the CF schema, though with more weight towards no change. 

\end{document}